\documentclass[letterpaper, 10 pt, conference]{ieeeconf}  
\usepackage{epsfig}
\usepackage{amstext}
\usepackage{amsmath}
\usepackage[latin1]{inputenc}  
\usepackage{amsfonts}
\usepackage{amssymb}
\usepackage{epstopdf}

\usepackage{fancyhdr}
\usepackage{textcomp}
\usepackage{float}
\usepackage{caption}
\usepackage{subcaption}
\usepackage{url} 
\usepackage{hyphenat}
\usepackage{hyperref}

\usepackage{algorithm}
\usepackage{algorithmic}

\usepackage{todonotes}
\usepackage{graphicx}
\usepackage[keeplastbox]{flushend}

\usepackage{color}

\IEEEoverridecommandlockouts                              

\title{Deep Lidar CNN to Understand the Dynamics of Moving Vehicles}

\author{Victor Vaquero, Alberto Sanfeliu, Francesc Moreno-Noguer
\thanks{The authors are with the Institut de Rob\`otica i Inform\`atica
Industrial, CSIC-UPC, Llorens i Artigas 4-6, 08028 Barcelona, Spain. {\tt \small
\{vvaquero,sanfeliu,fmoreno\}@iri.upc.edu}}%
\thanks{
This work has been supported by the Spanish Ministry of Economy and Competitiveness projects HuMoUR (TIN2017-90086-R) and COLROBTRANSP (DPI2016-78957-R) and
the Spanish State Research Agency through the Mar\'ia de Maeztu Seal of Excellence to IRI (MDM-2016-0656).
The authors also thank Nvidia for hardware donation under the GPU grant program.
}%
}

\begin{document}

\maketitle
\thispagestyle{empty}
\pagestyle{empty}

\begin{abstract}
Perception technologies in Autonomous Driving are experiencing their golden age due to the advances in Deep Learning. Yet, most of these systems rely on the semantically rich information of RGB images. Deep Learning solutions applied to  the data of other sensors typically mounted on autonomous cars (e.g. lidars or radars) are not explored much. 
In this paper we propose a novel solution to understand the dynamics of moving vehicles of the scene from only lidar information. 
The main challenge of this problem stems from the fact that we need to disambiguate the proprio-motion of the ``observer" vehicle from that of the external ``observed" vehicles.  For this purpose, we devise a CNN architecture which at testing time is fed with pairs of consecutive lidar scans.  
However, in order to properly learn the parameters of this network, during training we introduce a series of so-called pretext tasks which also leverage on image data. These tasks include  semantic information about vehicleness and a novel lidar-flow feature which combines standard image-based optical flow with lidar scans. We obtain very promising results and show that including distilled image information only during training, allows improving the inference results of the network at test time, even when image data is no longer used.

\end{abstract}



\section{INTRODUCTION} \label{Sec1_Intro}

Capturing and understanding the dynamics of a scene is a paramount ingredient for multiple autonomous driving (AD) applications such as obstacle avoidance, map localization and refinement or vehicle tracking. In order to efficiently and safely navigate in our unconstrained and highly changing urban environments, autonomous vehicles require precise information about the  semantic and  motion characteristics of the objects in the scene. Special attention should be paid to moving elements, e.g. vehicles/pedestrians, mainly if their motion paths are expected to cross the direction of other objects or the observer.

\begin{figure}[t]
\centering
 	\includegraphics[width=0.98    \columnwidth]{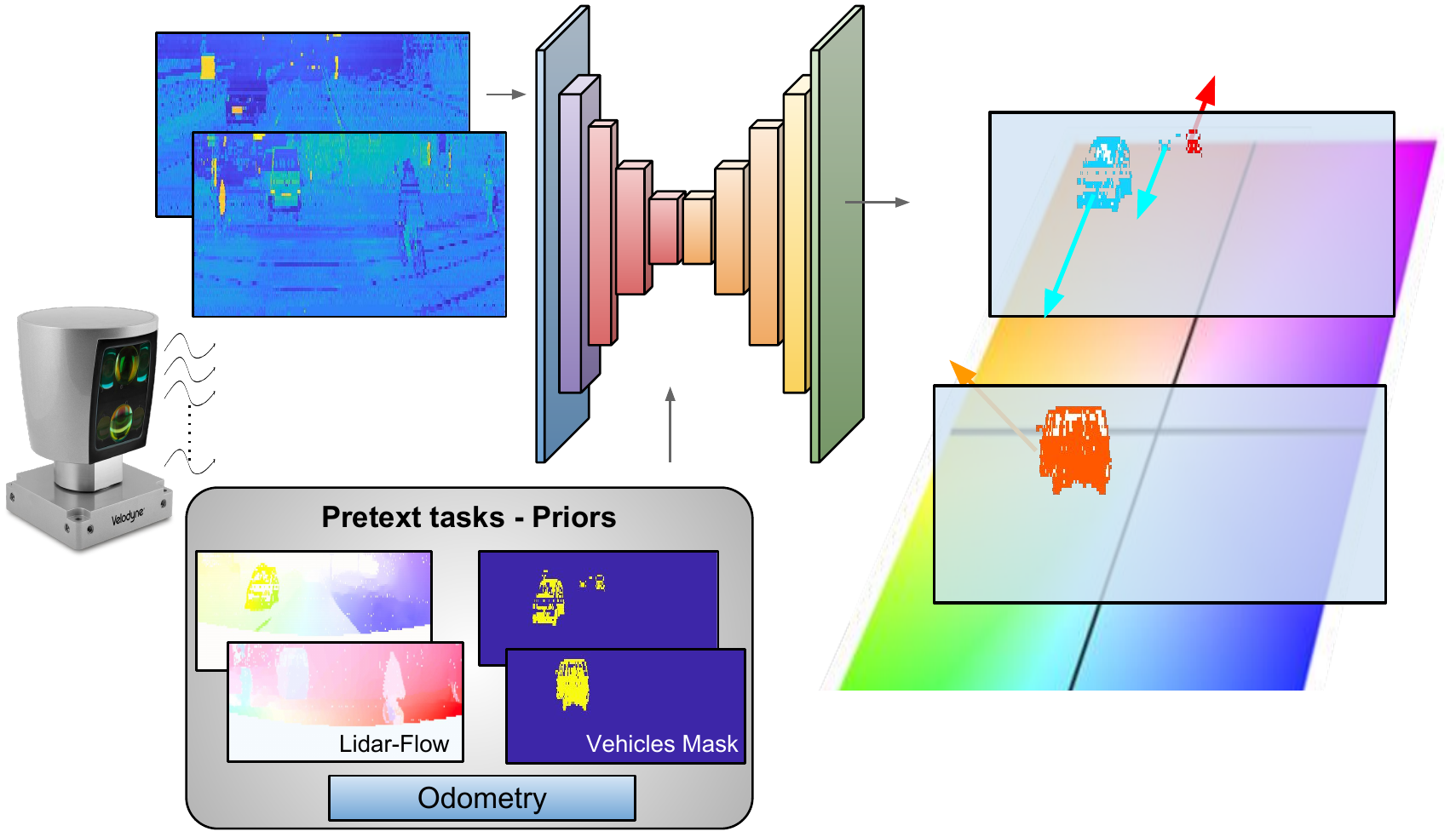}
 \caption{We present a deep learning approach that, using only lidar information, is able to estimate the ground-plane motion of the surrounding vehicles. In order to guide the learning process we introduce to our deep framework prior semantic and pixel-wise motion information, obtained from solving simpler pretext tasks, as well as odometry measurements.}
 \label{fig:caf_diagram}
\end{figure}

Estimating the dynamics of  moving objects in the environment requires both from advanced acquisition devices and interpretation algorithms.  
In autonomous vehicle technologies, environment information is captured through several sensors including cameras, radars and/or lidars. 
On distilling this data, research over RGB images from cameras has greatly advanced with the recent establishment of deep learning technologies. Classical perception problems such as semantic segmentation, object detection, or optical flow prediction~\cite{bai2016exploiting,geiger2012we,behl_iccv2017}, have experienced a great boost due to new Convolutional Neural Networks (CNNs) and their ability to capture complex abstract concepts given enough training samples. 
However, in an AD environment cameras may suffer a significant performance decrease under abrupt changes of illumination or harsh weather conditions, as for example driving at sunset, night or under heavy rain. 
On the contrary, radar and lidar-based systems performance is robust to these situations. 
While radars are able to provide motion clues of the scene,  their unstructured information and their lack of geometry comprehension make it difficult to use them for other purposes. 
Instead, lidar sensors provide very rich geometrical information of the 3D environment, commonly defined in a well-structured way.

In this paper we propose a novel approach to detect the motion vector of dynamic vehicles over the ground plane by {\em using only lidar information}. 
Detecting independent dynamic objects reliably from a moving platform (ego vehicle) is an arduous task. The proprio-motion of the vehicle in which the lidar sensor is mounted needs to be disambiguated from the actual motion of the other objects in the scene, which introduces additional difficulties.

We tackle this challenging problem by designing a novel Deep Learning framework. Given two consecutive lidar scans acquired from a moving vehicle, our approach is able to detect the movement of the other vehicles in the scene which have an actual motion with respect to a ``ground" fixed reference frame (see Figure~\ref{fig:caf_diagram}). 
During inference, our network is only fed with lidar data, although for training we consider a series of pretext tasks to help with solving the problem that can potentially exploit image information. 
Specifically, we introduce a novel \textit{lidar-flow} feature that is learned by combining lidar and standard image-based optical flow. 
In addition, we incorporate semantic vehicleness information from another network trained on singe lidar scans. 
Apart from these priors, we introduce knowledge about the ego motion by providing odometry measurements as inputs too. 
A sketch of our developed approach is shown in Figure~\ref{fig:caf_diagram}, where two different scenes are presented along with the corresponding priors obtained from the pretext tasks used. The final output shows the predicted motion vectors for each scene, encoded locally for each vehicle according to the color pattern represented in the ground.

An ablation study with several combinations of the aforementioned pretext tasks shows that the use of the lidar-flow feature throws very promising results towards achieving the overall goal of understanding the motion of dynamic objects from lidar data only.


\section{RELATED WORK} \label{Sec3_RFs}

Research on classical perception problems have experienced a great boost in recent years, mainly due to the introduction of deep learning techniques. Algorithms making use of Convolutional Neural Networks (CNNs) and variants have recently matched and even surpassed previous state of the art in computer vision tasks such as image classification, object detection, semantic segmentation or optical flow prediction \cite{he2016deep, ren2015faster, long2015fully, ilg2016flownet}.

However, the crucial problem of distinguishing if an object is moving disjointly from the ego motion remains challenging. 
Analysing the motion of the scene through RGB images is also a defiant problem recently tackled with CNNs, with several recent articles sharing ideas with our approach. In \cite{tokmakov2017learning}, the authors train a CNN network on synthetic data that taking as input the optical flow between two consecutive images, is able to mask independently moving objects. In this paper we  go a step further and not just distinguish moving objects from static ones, but also estimate their motion vector on the ground plane reference. Other methods try to disengage ego and real objects movement by inverting the problem. For instance, \cite{agrawal2015learning} demonstrate that a CNN trained to predict odometry of the ego vehicle, compares favourably to standard class-label trained networks on further trained tasks such as scene and object recognition. This fact suggests  that it is possible to exploit ego odometry knowledge to guide a CNN on the task of disambiguating our movement from the free scene motion, which we do in Section~\ref{sec:pretext}.

The aforementioned works, though, are not focused on AD applications. On this setting, common approaches   segment object motion by minimizing geometrically-grounded energy functions. \cite{menze2015object} assumes that outdoor scenes can decompose into a small number of independent rigid motions and jointly estimate them  by optimizing a discrete-continuous CRF. \cite{kao2016moving} estimates the 3D dynamic points in the scene through a vanishing point analysis of 2D optical-flow vectors. Then, a three-term energy function is minimized in order to segment the scene into different motions.

Lidar based approaches to solve the vehicle motion segmentation problem, have been led by clustering methods, either motion-  or model-based. The former~\cite{dewan2016motion}, estimates point motion features by means of RANSAC or similar methods, which then are  clustered to help on reasoning at object level. Model-based approaches, e.g.~\cite{vaquero_ecmr2017}, initially cluster vehicle points and then retrieve those which are moving by matching them through frames. 

Although not yet very extended, deep learning techniques are nowadays being also applied to the vehicle detection task over lidar information. 
\cite{li20163d} directly applies   3D convolutions over the point cloud euclidean space to detect and obtain the bounding box of vehicles. As these approaches are computationally very demanding, some authors try to alleviate this computational burden by sparsifying the convolutions over the point-cloud~\cite{engelcke2017vote3deep}. But still the main attitude is to project the 3D point cloud into a featured 2D representation and therefore being able to apply the well known 2D convolutional techniques~\cite{li2016vehicle, vaquero_ecmr2017}. In this line of projecting point clouds, other works propose to fuse of RGB images with lidar front and bird eye features~\cite{chen2016multi}.

However, none of these approaches is able to estimate the movement of the vehicles in an end-to-end manner without further post-processing the output as we propose. As far as we know, the closer work is~\cite{dewan2017deep} which makes use of RigidFlow~\cite{dewan2016rigid}  to classify each point as \textit{non-movable}, \textit{movable}, and \textit{dynamic}. In this work, we go a step further, and not only classify the dynamics of the scene, but also predict the motion vector of the  moving vehicles. 

Our approach also draws inspiration from progressive neural networks~\cite{rusu2016progressive} and transfer learning concepts~\cite{yosinski2014transferable} in that we aim to help the network to solve a  complex problem by solving a set of  intermediate ``pretext" tasks. For instance, in the problem of visual optical flow, \cite{sevilla2016optical} and~\cite{bai2016exploiting} use semantic segmentation pretext tasks. Similarly, during training, we also feed the network with prior knowledge about segmented vehicles on the point cloud information.


\section{Deep Lidar-based Motion Estimation} \label{Sec:Approach}

We next describe the proposed deep learning   framework to estimate the actual motion of vehicles in a scene independently from the ego movement, and  using only lidar information. 
Our approach relies on a Fully Convolutional Network that receives as input featured lidar information from two different but temporary close frames of the scene, and outputs a dense estimation of the ground-floor motion vector of each point, given the case that it belongs to a dynamic vehicle. 
For that, in Section~\ref{sec:imdb} we introduce a novel dataset built from the Kitti tracking benchmark that has been specifically created to be used as ground-truth in our supervised problem. 
Since lidar information by itself is not enough to solve the proposed complex mission, in Section~\ref{sec:pretext} we consider exploiting pretext tasks to introduce prior knowledge about the semantics and the dynamics of the scene to the main CNN defined in Section~\ref{sec:net}.

\begin{figure*}[t!]
\centering
\begin{subfigure}{.5\textwidth}
  \centering 
  \includegraphics[width=0.98    \columnwidth]{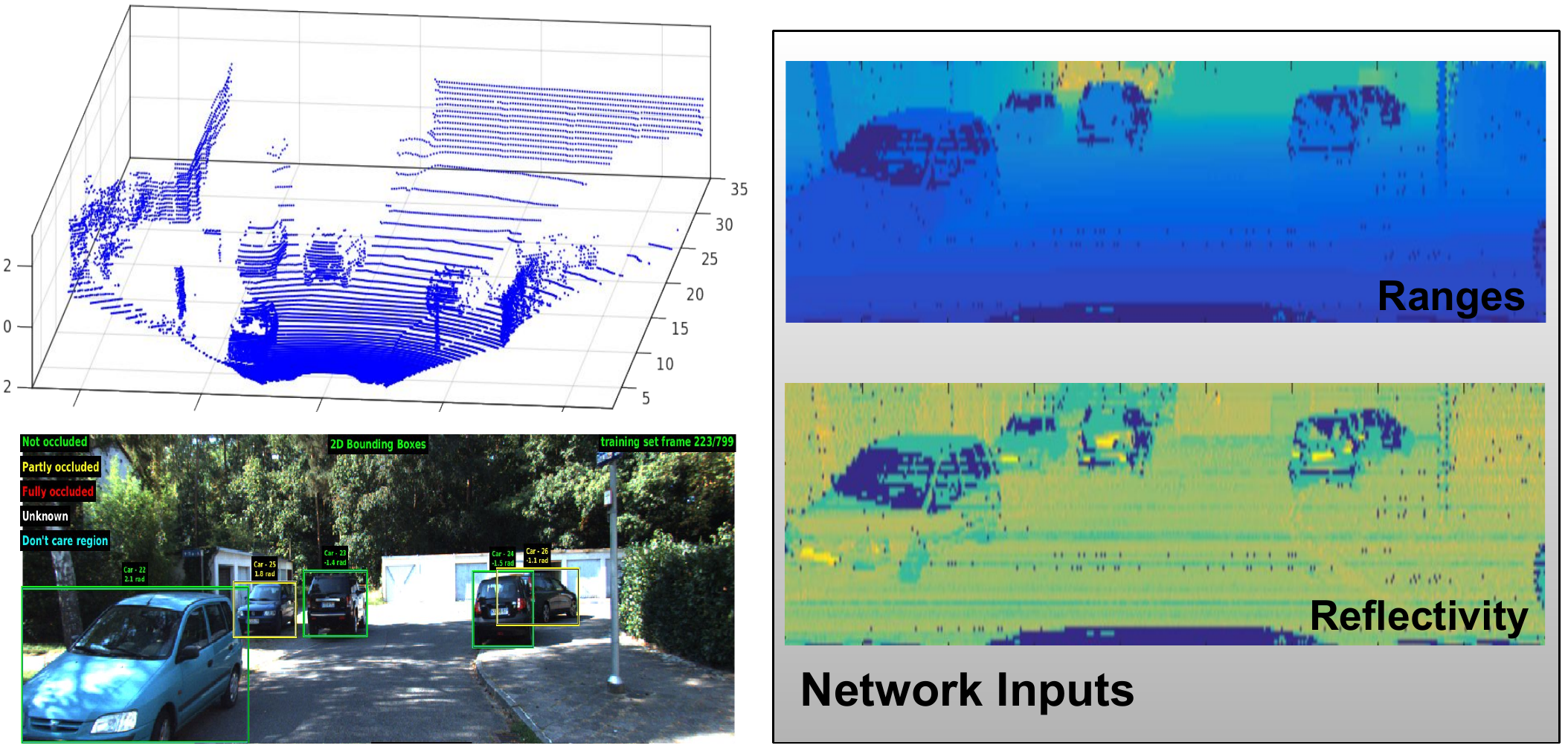}	
  \caption{Basic point cloud frame input data. Ranges and reflectivity}
  \label{fig:input_data}
\end{subfigure}%
\begin{subfigure}{.5\textwidth}
  \centering
  \includegraphics[width=0.97    \columnwidth]{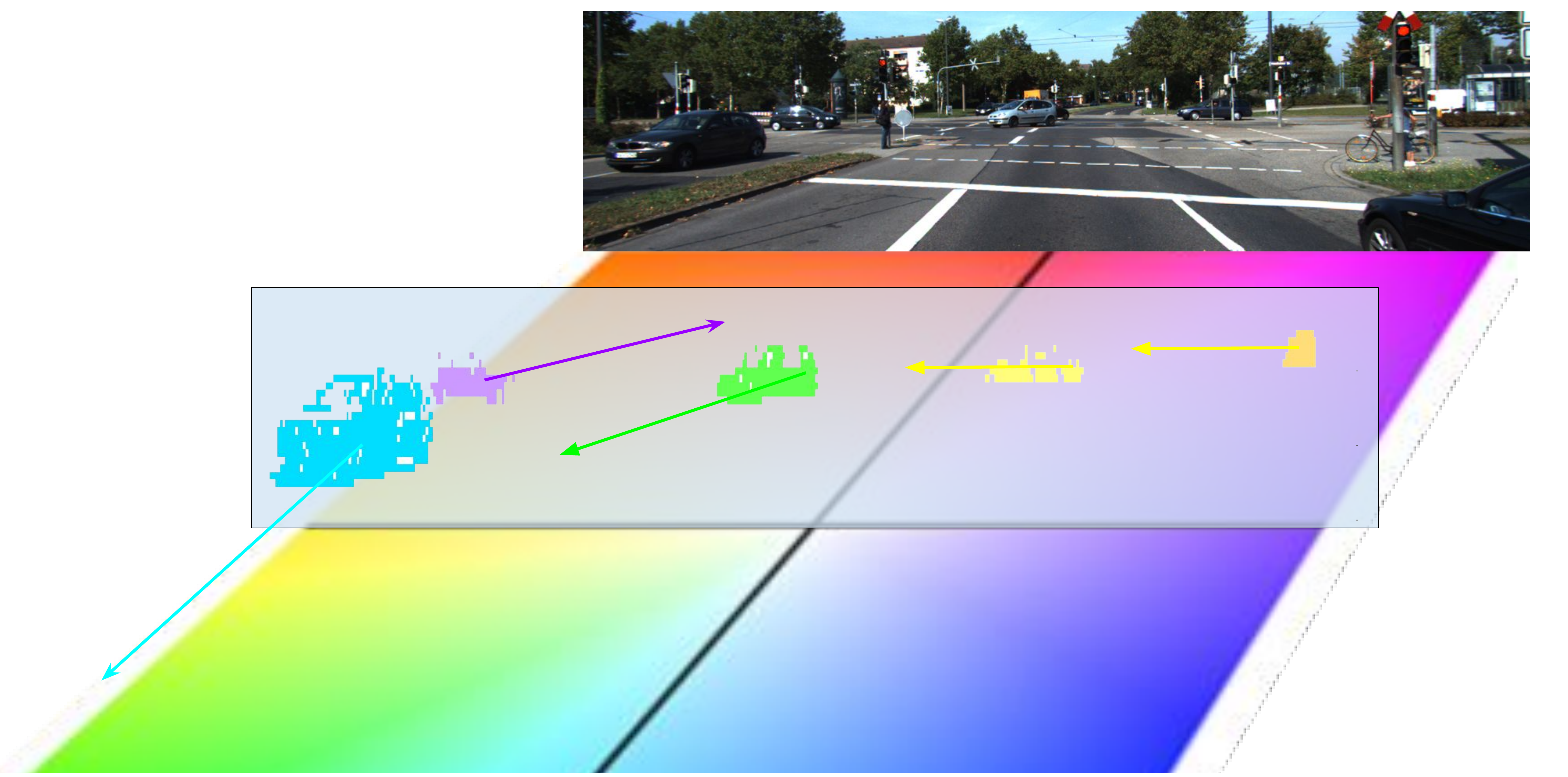}	
  \caption{Network output: lidar-based vehicle motion representation.}
  \label{fig:output_data}
\end{subfigure}
\caption{{Basic input and predicted output. (a) The input of the network corresponds to a pair of lidar scans, which are represented in 2D domains of range and reflectivity. (b) The output we seek to learn represents the motion vectors of the moving vehicles, colored here with the shown pattern attending to the motion angle and magnitude.}}
\label{fig:input_output}
\end{figure*}

\subsection{Lidar Motion Dataset} \label{sec:imdb}

In order to train a CNN in a supervised manner, we need to define both the input information and the ground-truth of the desired output from which to compare the learned estimations. 

The simpler input data we use consists on the concatenation of two different projected lidar scans featuring the ranges and reflectivity measured, as the one shown in Figure~\ref{fig:input_data}.  
For each scan, we transform the corresponding points of the point cloud from its 3D euclidean coordinates to spherical ones, and project those to a 2D domain attending to the elevation and azimuth angles. Each row  corresponds to one of the sensor vertical lasers ($64$) and each column is equivalent to a laser step in the horizontal field of view ($448$). This field of view is defined attending to the area for which the Kitti dataset provides bounding box annotations, that covers approximately a range of $[-40.5, 40.5]$ degrees from the ego point of view. Each pair $(u,v)$ of the resulting projection encodes the sensor measured range and reflectivity.
A more detailed description of this process can be found in~\cite{vaquero_ecmr2017}.

To build the desired ground-truth output, we make use of the annotated 3D bounding boxes of the Kitti Tracking Benchmark dataset \cite{geiger2012we}, that provides diversity of motion samples. 
As nowadays vehicles still move on the ground plane, we stated its motion as a two~-~dimensional vector over the Z/X plane, being Z our forward ego direction and X its transversal one. 
For each time $t$, we define our ego-vehicle position as $O_t \in \mathbb{R}^2$, i.e \textit{the observer}. 
Considering there are $\mathcal{X}$ vehicle tracks in the scene at each moment, we define any of these vehicle centroids as seen from \textit{the observer} frame of reference like $^{O_{t}}C_{t,x} \in \mathbb{R}^2$ where $x = 1 \dots \mathcal{X}$. For a clearer notation, we will show a use case with just one free moving vehicle, omitting therefore the $x$ index.
As both \textit{the observer} and the other vehicle are moving, we will see the free vehicle centroid $^{O_{t+n}}C_{t+n}$ each time from a different position $O_{t+n}$.
Therefore, in order to get the real displacement of the object in the interval $t \rightarrow t+n$ we need to transform this last measurement to our previous reference frame $O_t$, obtaining $^{O_{t}}C_{t+n}$. 
Let us denote our own frame displacement as $^{O_t}T^{t+n}_{t}$, which is known by the differential ego-odometry measurements. Then, the transformed position of the vehicle centroid is simply:
\begin{equation}
	^{O_{t}}C_{t+n} = (^{O_t}T^{t+n}_{t})^{-1} \,\cdot\, ^{O_{t+n}}C_{t+n}
\end{equation}
and the on-ground motion vector of the free moving vehicle in the analysed interval can be calculated as $^{O_{t}}C_{t+n} \, - \, ^{O_{t}}C_{t}$.

Notice that these ground-truth needs to be calculated in a temporal sliding window manner using the lidar scans from frames $t$ and $t+n$ and therefore, different results will be obtained depending on the time step $n$. The bigger this time step is, the longer will be the motion vector, but it will be harder to obtain matches between vehicles. 

Some drift is introduced as accumulation of errors from i) the Kitti manual annotation of bounding boxes, ii) noise in the odometry measurements and iii) the transformations numerical resolution. This made that some static vehicles were tagged as slightly moving. 
We therefore filtered the moving vehicles setting as dynamic only the ones which displacement is larger than a threshold depending on the time interval $n$, and consequently, directly related to the minimum velocity from which we consider a movement. We experimentally set this threshold to 10Km/h.
 
Finally, we encode each vehicle ground-truth motion vector attending to its angle and magnitude according to the color-code typically used to represent optical flow. Figure~\ref{fig:output_data} shows a frame sample of the described dataset, where the corresponding RGB image of the scene is also shown just for comparison purposes. \footnote{We plan to make our datasets publicly available.}.

 \begin{figure*}[t!]
\centering

\begin{subfigure}{.5\textwidth}
  \centering
  \includegraphics[width=0.98    \columnwidth]{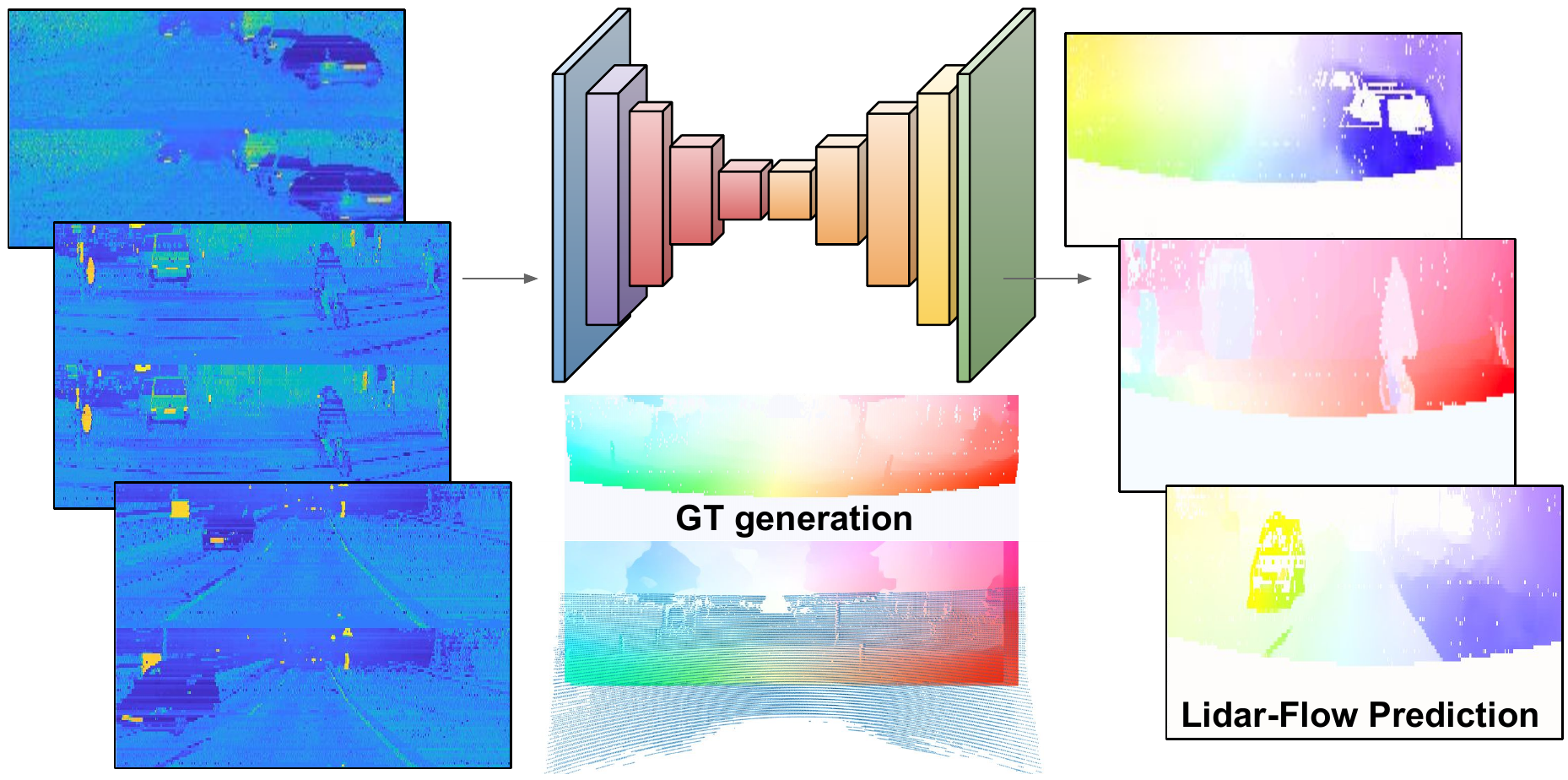}
  \caption{{Lidar-flow prior}}
  \label{fig:motion_prior}
\end{subfigure}%
\begin{subfigure}{.5\textwidth}
  \centering
  \includegraphics[width=0.95    \columnwidth]{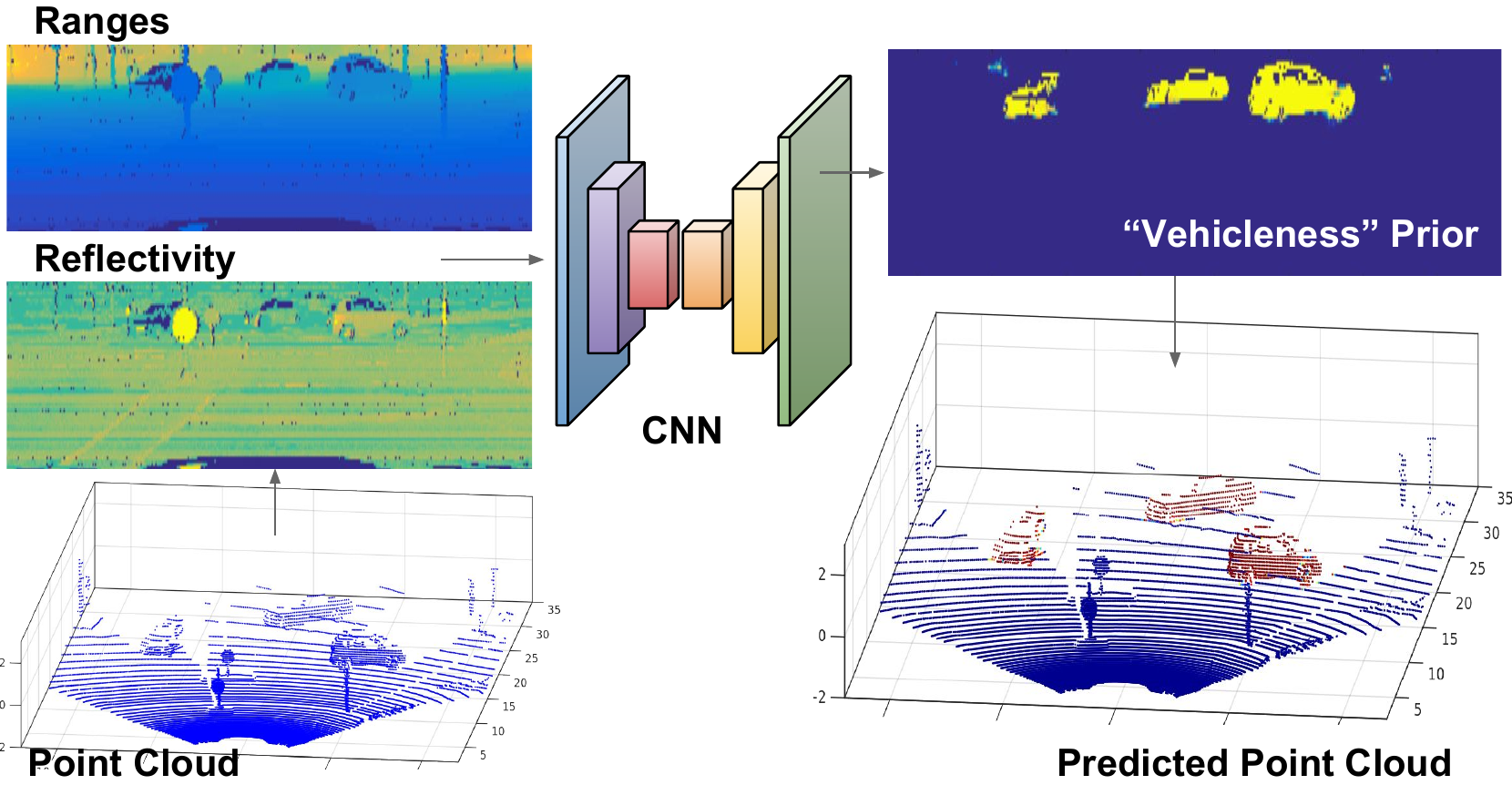}
  \caption{{Semantic prior}}
  \label{fig:semantic_prior}
\end{subfigure}
\caption{{ Pretext tasks used to guide the final learning. a) lidar-flow prior, obtained by processing pairs of frames through a new learned lidar-flow net; b) semantic prior, obtained by processing single frames through our vehicle lidar-detector net.}}
\label{fig:priors}
\end{figure*}

\subsection{Pretext tasks} \label{sec:pretext} 

As aforementioned, we guide the network learning towards the correct solution introducing prior knowledge obtained by solving other pretext tasks. This idea draws similarities from progressive networks \cite{rusu2016progressive} and transfer learning works \cite{agrawal2015learning}, both helping in solving increasing complexity tasks.
In this manner, we introduce three kinds of additional information: a) a lidar-optical flow motion prior to guide our network for finding matches between the two lidar inputs; b) semantic concepts that will help with focusing on the vehicles in the scene; c) the ego motion information based on the displacement given by the odometry measurements.

The motion prior for matching inputs is given by stating a novel deep-flow feature map that can be seen in Figure~\ref{fig:motion_prior}. 
We developed a new deep framework that takes as input the 2D projection of lidar scans from two separated frames and outputs a learned lidar-based optical flow.
As lidar-domain optical flow ground-truth is not available, we created our own for this task. To do this, we used a recent optical flow estimator \cite{ilg2016flownet} based on RGB images and obtained flow predictions for the full Kitti tracking dataset. We further created a geometric model of the given lidar sensor attending to the manufacturer specifications and projected it over the predicted flow maps, obtaining the corresponding lidar-flow of each point in the point cloud. Finally we stated a dense deep regression problem that uses the new lidar-flow features as ground-truth to learn similar 2D motion patterns using a flownet-alike convolutional neural network. 

Semantic priors about the vehicleness of the scene are introduced via solving a per-pixel classification problem like the one presented in~\cite{vaquero_ecmr2017}. For it, a fully convolutional network is trained to take a lidar scan frame and classifies each corresponding point as belonging to a vehicle or background. An example of these predictions is shown in Figure \ref{fig:semantic_prior}. 

Finally, we introduce further information about the ego-motion in the interval. For this, we create a 3 channel matrix with the same size as the 2D lidar feature maps where each ``pixel'' triplet takes the values for the forward (Z) and transversal (X) ego-displacement as well as the rotation over Y axis in the interval $t \rightarrow t+n$.

\subsection{Deep-Lidar Motion Network} \label{sec:net}

\begin{figure*}[t]
\centering
  	\includegraphics[width=0.95\linewidth]{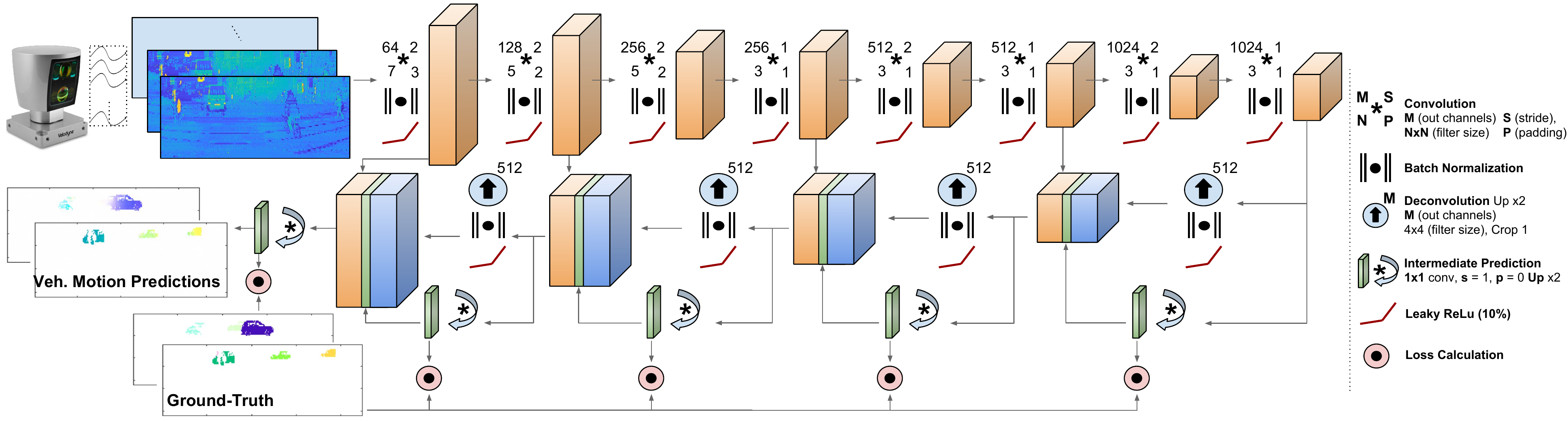}
	\caption{{Deep Learning architecture used to predict the motion vector of moving vehicles from two lidar scans.}}
	\label{fig:net}
\end{figure*}

As network architecture for estimating the rigid motion of each vehicle over the ground floor, we considered the Fully Convolutional Network detailed in Figure \ref{fig:net}. It draws inspiration from FlowNet \cite{dosovitskiy2015flownet}, which is designed to solve a similar regression problem. However, we introduced some changes to further exploit the geometrical nature of our lidar information. 

We first transformed the network expansive part by introducing new deconvolutional blocks at the end with the respective batch normalization (BN) and non-linearity imposition (Relu). Standard FlowNet output is sized a fourth of the input and bi-linearly interpolated in a subsequent step. This is not applicable to our approach as our desired output is already very sparse containing only few groups of lidar points that belong to moving vehicles. Therefore mid resolution outputs may not account for far vehicles that are detected by only small sets of points.
In addition, we eliminate both the last convolution and first deconvolution blocks of the inner part of FlowNet, for which the generated feature maps reach a resolution of $1/64$ over the initial input size. Note that our lidar input data has per se low resolution ($64\times448$), 
and performing such an aggressive resolution reduction has been shown to result in missing targets. 
On the other hand, we follow other FlowNet original attributes. Thus, our architecture performs a concatenation between equally sized feature maps from the contractive and the expansive parts of the network which produce richer representations and allows better gradient flow. In addition, we also impose intermediate loss optimization points obtaining results at different resolutions which are upsampled and concatenated to the immediate upper feature maps, guiding the final solution from early steps and allowing the back-propagation of stronger and healthier gradients. 

\section{Experiments} \label{Sec:exps}

This section provides a thorough analysis of the performance of our framework fulfilling the task of estimating the motion vector of moving vehicles given two lidar scans.

\subsection{Training details}

For training the presented deep neural networks from both the main framework and the pretext tasks, we set $n$ to $1$, so that measuring the vehicles movement between two time consecutive frames.
All these networks are trained from zero, initializing them with the He's method~\cite{he2015delving} and using Adam optimization with the standard parameters $\beta_1 = 0.9$ and $\beta_2 = 0.999$. 

The Kitti Tracking benchmark contains a large number of frames with static vehicles, which results in a reduction of the number of samples from which we can learn. 
Our distilled Kitti Lidar-Motion dataset contains $4953$ frames with moving vehicles, and $3047$ that either contain static vehicles or do not contain any. To balance the batch sampling and avoid a biased learning, we take for each batch $8$ frames containing movement and $2$ that do not. For training all models we left for validation the sequences $1$, $2$ and $3$ of our distilled Kitti Training dataset, which results in $472$ samples with motion and $354$ without. 
As our training samples represent driving scenes, we perform data augmentation providing only horizontal flips with a $50\%$ chance, in order to preserve the strong geometric lidar properties.

The training process is performed on a single NVIDIA 1080 Ti GPU for $400,000$ iterations with a batch size of $10$ Velodyne scan pairs per iteration. The learning rate is fixed to $10^{-3}$ during the first $150,000$ iterations, after which, it is halved each $60,000$ iterations. 
As loss, we use the euclidean distance between the ground-truth and the estimated motion vectors for each pixel. We set all the intermediate calculated losses to equally contribute for the learning purposes.

\subsection{Input data management for prior information}

Depending on the data and pretext information introduced in our main network, different models have been trained; following, we provide a brief description.

Our basic approach takes as input a tensor of size $64 \times 448 \times 4$ which stacks the 2D lidar projected frames from instants $t$ and $t+1$. Each projected frame contains values of the ranges and reflectivity measurements, as summarized at the beginning of Section \ref{sec:imdb} and shown in Figure \ref{fig:input_data}.

For obtaining motion priors, the lidar data is processed through our specific lidar-flow network. It produces, as output, a two channel flow map where each pair \textit{(u,v)} represents the RGB equivalent motion vector on the virtual camera alike plane as shown in Figure \ref{fig:motion_prior}.  
When incorporating this motion prior to the main network, the lidar-flow map is concatenated with the basic lidar input to build a new input tensor   containing $6$ depth channels.

In order to add semantic prior knowledge in the training, we separately process both lidar input frames through our learned vehicle detection network \cite{vaquero_ecmr2017}. The obtained outputs indicate the predicted probability of each pixel to belong to a vehicle. This information is further concatenated with the raw lidar input plus the lidar-flow maps, yielding a tensor with a depth of $8$ channels.

Finally, for introducing the odometry information as well, three more channels are concatenated to the stacked input resulting in a tensor of depth $11$.

\subsection{Results} \label{Sec:results}

Table \ref{tab:results} shows a quantitative analysis of our approach. We demonstrate the correct performance of our framework by setting two different baselines, \textit{error@zero} and \textit{error@mean}. The first one assumes a \textit{zero} regression, so that sets all the predictions to zero as if there were no detector. The second baseline measures the end-point-error that a mean-motion output would obtain.

Notice that in our dataset only a few lidar points fall into moving vehicles on each frame. Therefore, measuring the predicted error over the full image does not give us a notion about the accuracy of the prediction, as errors generated by false negatively (i.e. that are dynamic but considered as static without assigning them a motion vector) and false positively (i.e. that are static but considered as dynamic assigning them a motion vector) detected vehicles, would get diluted over the full image. In order to account for this fact, we also measure end-point error over the real dynamic points only. Both measurements are indicated in Table \ref{tab:results} as \textit{full} and \textit{dynamic}. All the given values are calculated at test time over the validation set only, which during the learning phase has never been used for training neither the main network nor the pretext tasks. Recall that during testing, the final networks are evaluated only using lidar Data.

We tagged the previously described combinations of inputs data with $D$, $F$, $S$ and $O$ respectively for models using \textit{Data}, \textit{Lidar-Flow}, \textit{Vehicle Segmentation} and \textit{Odometry}. When combining different inputs, we express it with the $\&$ symbol between names; e.g. a model named $D$  $\&$  $F$  $\&$  $S$ has been trained using as input the lidar \textit{Data}, plus the priors \textit{Lidar-Flow} and \textit{Vehicle Segmentation}.

\begin{table}[]
\centering
\caption{Evaluation results after training the network using several combination of lidar- and image-based priors. Test is performed using only lidar inputs.  Errors are measured in pixels, end-point-error.}
\label{tab:results}
\begin{tabular}{l||c|c||}
\multicolumn{1}{l||}{} & \multicolumn{1}{c|}{full} & \multicolumn{1}{c||}{dynamic} \\
 \hline \hline
error@zero  				&   0.0287    	&    1.3365      \\ 
error@mean  				&   0.4486    	&    1.5459      \\ \hline
D						&   0.0369		&	 1.2181		\\
GT.F      				&   0.0330    	&    1.0558		\\ 
GT.F \& D \& S			&   0.0234    	&    0.8282		\\ \hline
Pred.F      				&   0.0352    	&    1.1570		\\ 
Pred.F \& D 				&   0.0326   	&    1.1736   	     \\ 
Pred.F \& D \& S 		&   0.0302    	&    1.0360         \\ 
Pred.F \& D \& S \& O 	&   0.0276    	&    0.9951   	   \\ 
\hline

\end{tabular}
\end{table}

At the light of the results several conclusions can be extracted. 
First, is that only lidar information is not enough for solving the problem of estimating the motion vector of freely moving vehicles in the ground, as we can see how the measured dynamic error is close to the error at zero.  
To account for the strength of introducing optical flow as motion knowledge for the network, we tested training with only the lidar-flow ground truth (GT.F rows in the table) as well as with a combination of flow ground-truth, semantics and lidar data. Both experiments show favourable results, being the second one the most remarkable. 
However, the lidar-flow ground-truth is obtained from the optical flow extracted using RGB images, which does not accomplish our \textit{solo-lidar} goal. We therefore perform the rest of experiments with the learned lidar-flow (Pred.F rows in the table) as prior, eliminating any dependence on camera images. As expected, our learned lidar-flow introduce further noise but still allow us to get better results than using only lidar information, which suggest that flow notion is quite important in order to solve the major task.


\section{Conclusions} \label{Sec5_conclusions}
In this paper we have addressed the problem of understanding the dynamics of moving vehicles from lidar data acquired by a vehicle which is also moving. Disambiguating proprio-motion from other vehicles' motion poses a very challenging problem, which we have tackled using  Deep Learning. The main contribution of the paper has been to show that while during testing, the proposed Deep Neural Network is only fed with lidar-scans, its performance can be boosted by exploiting other prior image information during training. We have introduced a series of pretext tasks for this purpose, including semantics about the vehicleness and an optical flow texture built from both image and lidar data. The results we have reported are very promising and demonstrate that exploiting image information only during training really helps the lidar-based deep architecture. In future work, we plan to further exploit this fact by introducing other image-based priors during training, such as the semantic information of all object categories in the scene and dense depth obtained from images.

\bibliographystyle{ieeetr}
\bibliography{icra18_DLMF_bib}

\end{document}